%% file: main.tex
\documentclass[conference]{IEEEtran}
\input{macros}

\begin{document}

\title{A Generative Approach for Production-Aware Industrial Network Traffic Modeling}

\input{authors}

\maketitle

\begin{abstract}
The new wave of digitization induced by Industry 4.0 calls for ubiquitous and reliable connectivity to perform and automate industrial operations. 5G networks can afford the extreme requirements of heterogeneous vertical applications, but the lack of real data and realistic traffic statistics poses many challenges for the optimization and configuration of the network for industrial environments. In this paper, we investigate the network traffic data generated from a laser cutting machine deployed in a Trumpf factory in Germany. We analyze the traffic statistics, capture the dependencies between the internal states of the machine, and model the network traffic as a production state dependent stochastic process. The two-step model is proposed as follows: first, we model the production process as a multi-state semi Markov process, then we learn the conditional distributions of the production state dependent packet interarrival time and packet size with generative models. We compare the performance of various generative models including variational autoencoder (VAE), conditional variational autoencoder (CVAE), and generative adversarial network (GAN). The numerical results show a good approximation of the traffic arrival statistics depending on the production state. Among all generative models, CVAE provides in general the best performance in terms of the smallest Kullback-Leibler divergence.
\end{abstract}



\section{Introduction}\label{sec:1}

\input{section1}
\section{Data collection and dataset description}\label{sec:2}
\input{section2}
\section{Modeling the Production-Aware Network Traffic Process}\label{sec:3}
\input{section3}

\section{Performance evaluation}\label{sec:4}
\input{section4}

\section{Conclusion}\label{sec:5}
\input{conclusion}
\section*{Acknowledgment}\label{Section:Acknowledgement}
\input{acknowledgement}

\bibliographystyle{IEEEtran}
\bibliography{bibliography}

\end{document}

%% file: macros.tex
\usepackage{url}
\usepackage{bbm}
\usepackage[utf8]{inputenc}
\usepackage{cite}
\usepackage{amsmath,amssymb,amsfonts,amsbsy}
\usepackage{dsfont}
\usepackage{caption}
\usepackage[labelformat=simple]{subcaption}

\usepackage{algpseudocode}
\usepackage{algorithm}
\usepackage{booktabs}
\usepackage{graphicx}
\usepackage{textcomp}
\usepackage{xcolor}
\usepackage{tikz}

\def\BibTeX{{\rm B\kern-.05em{\sc i\kern-.025em b}\kern-.08em
    T\kern-.1667em\lower.7ex\hbox{E}\kern-.125emX}}

\newcommand{\set}[1]{\mathcal{#1}}

\newcommand{\ma}[1]{\boldsymbol{\mathbf{#1}}}
\newcommand{\ve}[1]{\boldsymbol{\mathbf{#1}}}

\newcommand{\Ss}{{\set{S}}}
\newcommand{\mP}{{\ma{P}}}
\newcommand{\vx}{{\ve{x}}}

\newcommand{\vz}{{\ve{z}}}

%% file: authors.tex
\author{
    \IEEEauthorblockN{Alessandro~Lieto\IEEEauthorrefmark{1}, Qi~Liao\IEEEauthorrefmark{1}, Christian~Bauer\IEEEauthorrefmark{2}}
        \IEEEauthorblockA{\IEEEauthorrefmark{1}Nokia Bell Labs, Stuttgart, Germany  \\
        \IEEEauthorrefmark{2}Trumpf, Ditzingen, Germany
   \\ E-Mails: alessandro.lieto, qi.liao@nokia-bell-labs.com, christian.bauer@trumpf.com
}
}

%% file: section1.tex
The industrial network infrastructures are facing a radical transformation to accommodate the communication needs of today's manufacturing technologies, Industrial Internet of Things (IIoT) applications and flexible shopfloor designs. The current industrial networks are designed for static configuration and specific applications, and they are, therefore, insufficient to cover the future needs posed by Industry 4.0~\cite{rao2018impact}. 5G systems and beyond are a natural fit for serving the evolving needs of future industrial environments, by providing the dynamicity and flexibility that is expected to be supported in the near future~\cite{malanchini2021leveraging}. However, there is a lack of knowledge about the characteristics of industrial systems, and the typical configuration of cellular systems tailored to consumer applications is not able to meet the demanding needs of industrial applications. Especially for network planning, it is critical to know and quantify the amount of data to be transmitted, the statistics of data packets, the endpoints of the communication and many other details that require the direct observation from brownfield installations~\cite{lavassani2021:brownfield}. 

In general, the existing literature presents simplified findings from measurements performed over non-industrial networks or semi-artificial system setups. For example, there is a lot of literature focusing on the analysis and design of traffic in Supervisory Control and Data Acquisition (SCADA) networks, including datasets~\cite{scada:data} open to access for researchers. Among others,~\cite{lin2018:scada} analyzes the traffic patterns and models the interarrival time and the correlation between packet types generated in SCADA networks, limiting their observation to the industrial protocols applied therein. 
However, their evaluation is based on emulated traffic in test labs and not exemplary of realistic applications running on devices in a working industrial environment, i.e. not being able to address the complexity of the real industrial systems~\cite{lavassani2021:brownfield}.
Indeed, the validation of results with realistic assumptions is very challenging for researchers in this area due to the unavailability of both production and network traffic data from industrial systems. 
Some works, such as~\cite{lavassani2021:brownfield, Mogesen2021:models}, analyze the differences between available theoretical and empirical traffic models with real brownfield installations, showing that there is a significant gap in the literature, especially for traffic estimation and generation. They provide some interesting findings, but do not share data or models that can be used to guide the research in this field. While~\cite{lavassani2021:brownfield} mainly focuses on the methodology for best practice in traffic estimation and modeling, the authors in~\cite{Mogesen2021:models} provide an empirical comparison between the well-known analytical models suggested from standardization bodies, such as 3rd generation partnership project (3GPP) and the 5G-Alliance for Connected Industries and Automation (5G-ACIA)~\cite{3gpp2020, 5g-acia2019}, and real data traffic collected from various factories. Their analysis focuses on the statistics of interarrival packet time, packet size and burstiness, showing that the real traffic is significantly more heterogeneous than the suggested models from 3GPP, with only limited match concerning the periodic traffic, and with high over-estimation with respect to the non-periodic traffic. 
Along the same lines, the methodology proposed in~\cite{lavassani2022:traffic} considers the whole complexity of an industrial systems, when collecting all the aggregated traffic in an industrial network. 
Although this approach allows for higher complexity and gives insights for the design of a communication systems considering the overall network data traffic, it does not provide enough elements for repeatability. This is the most important point that we address in this paper. The contributions of the paper can be summarized as follows:
\begin{itemize}
	\item First, we analyze the peculiar behavior of some industrial devices from measurements collected in a Trumpf factory. Based on the operations of the machines, we can identify specific production dependent states of the machines and model their transitions as a semi-Markov process. 
	\item Then, we apply generative models for reproducing the statistics of data packets, focusing in particular on the interarrival time and its dependency on packet size.
	\item Finally, we compare different models that can be used for reproducing the traffic pattern of the machine for each of its production state. We also grant access to those models and share them publicly with the research community for validation and testing in various 5G use cases, e.g., to model a digital twin of an industrial network~\cite{git:models2022}. Although our models cannot generalize for any industrial environment, it still provides a reference point for realistic industrial installations and can be used to generate more realistic instances for e.g. industrial network simulators~\cite{kick2022}.
\end{itemize}

The rest of the paper is structured as follows. In Section~\ref{sec:2}, we describe the datasets and the characteristics of the traffic flows captured from laser cutting machines in a Trumpf factory. The methodology for the generative models is discussed in Section~\ref{sec:3}, while in Section~\ref{sec:4} we illustrate our findings and analyze the performance of the applied models. Finally, Section~\ref{sec:5} concludes the paper with a summary of the findings and discussion for the applications of the obtained models in industrial network setups.

%% file: section2.tex
We consider an exemplary industrial setup of a Trumpf factory where a laser cutting machine is deployed in a shop floor and communicates with a central management system over an industrial fixed network. All the information that have been exchanged between the machine and the management system are collected in a log file in .csv format, which represents the dataset of our evaluation. 
We filter the original dataset to capture only the entries needed for analyzing the statistical properties of the traffic generated by the industrial machine and to identify the specific characteristics of the machines, with its life-cycle, operational duty-cycle and status information to correlate the statistics of the traffic to the production state of the machine.
The filtered dataset contains the following list of entries.
\begin{itemize}
	\item \textit{Processed Time} identifies the date and time of the day when a specific message has been sent by the machine. The time granularity is of millisecond scale, which is appropriate for the estimation of packet interarrival time.
	\item \textit{Data.id} specifies the type of message sent by the machine. It is an ID used by the owner of the machine to recognize the messages that have been exchanged. We can use this information to identify control messages that are sent by the machine to inform about its current operational state.
	\item \textit{Data.value} represents the content of the message. The content of the message is out of the scope of our evaluation, however some information can be useful to identify specific status information of the machines. In particular, we extract together with the \textit{Data.id} the information that identifies what is the status of the machine. An example of how to use such information is described in the following subsection. 
	\item \textit{Data.payload} defines the size of each packet that has been transmitted by the device. The computation of the payload is rather simple, since we apply binary conversion to the content of \textit{Data.value} and we extract the dimensions in bytes. Although this calculation does not represent exactly the packet size over the network interface (since header and encryption are not considered in the calculation), it still provides coarse estimation of the amount of data to be transmitted. The payload has been then quantized in multiple of $32$ bytes to align with the typical minimum packet size of 3GPP standards.
\end{itemize}

%

\subsection{Production state transition of the laser cutting machine}

\begin{figure}[!t]
	\centering
	\includegraphics[width=\linewidth]{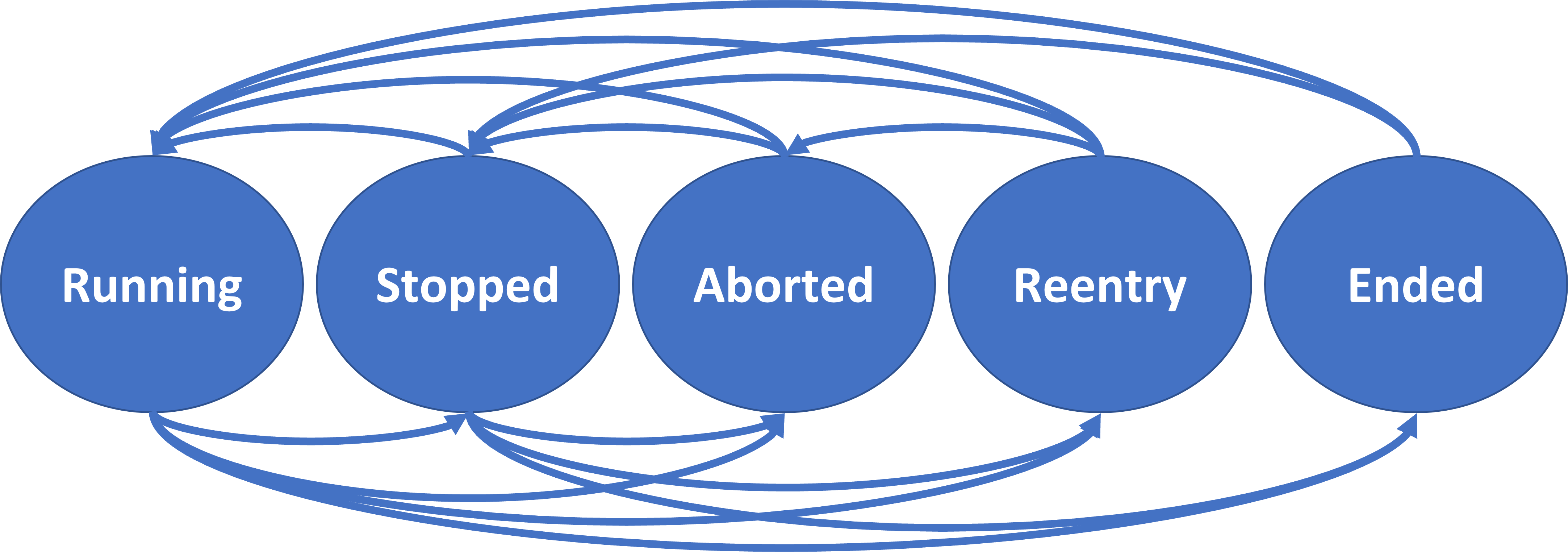}
	\caption{A Markov chain of state transition of the machine}
	\label{fig:markov}
	\vspace{-2ex}
\end{figure}

\begin{table}
\vspace{.4cm}
	\centering
	\begin{tabular}{ | c | c | c | c | c | c |}
		\hline
			  					& \textit{Running}       & \textit{Reentry}  	& \textit{Stopped}	& \textit{Aborted}	& \textit{Ended} \\ \hline
		\textit{Running}				& $0$   			& $4$			& $296$   			& $17$			& $151$		\\ \hline
		\textit{Reentry}  			& $51$      		& $0$			& $36$   			& $21$			& $0$	     	\\ \hline
		\textit{Stopped}        			& $198$ 	 		& $52$ 			& $0$ 	   		& $2$			& $15$		\\ \hline
		\textit{Aborted}        			& $9$  			& $0$ 			& $31$   			& $0$			& $0$	 	\\ \hline
		\textit{Ended}  				& $63$      		& $0$			& $103$   			& $0$			& $0$	     	\\ \hline
	\end{tabular}
	\caption{The number of the total transitions among states}
	\label{table:transition_state}
	\vspace{-.3cm}
\end{table}

\begin{figure}[!t]
	\centering
	\includegraphics[width=\linewidth]{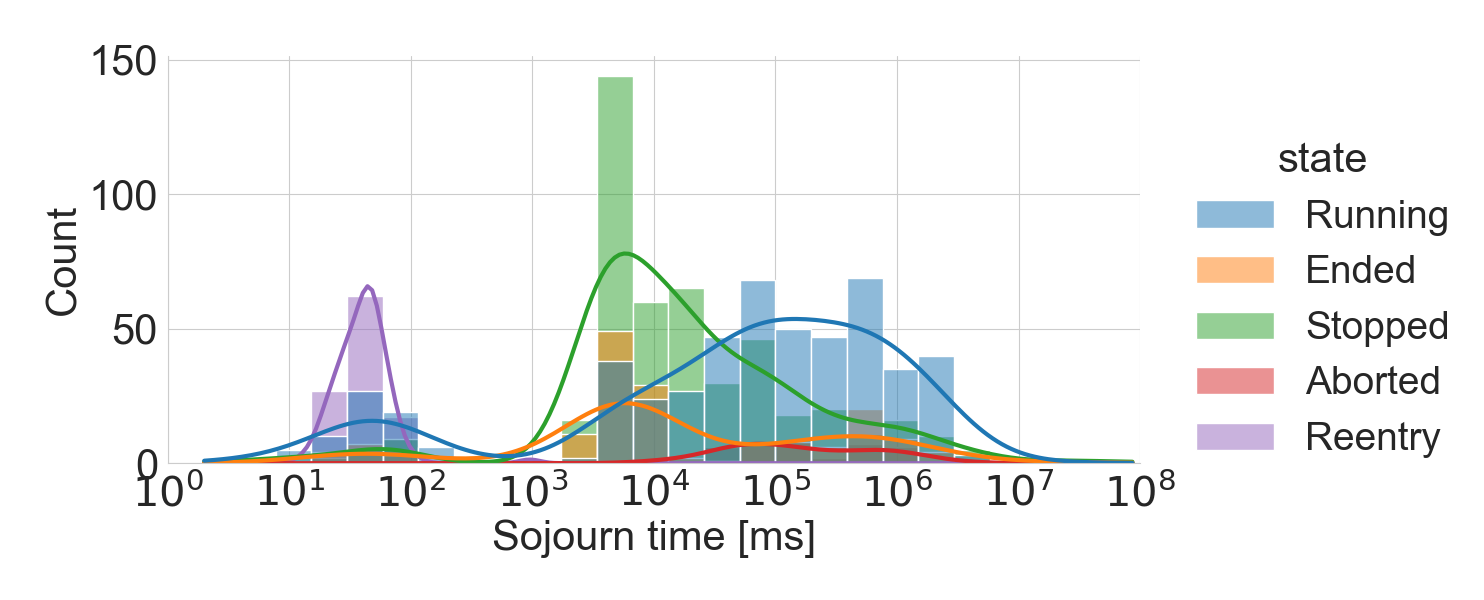}
	\caption{The distribution of sojourn time of the machine per state}
	\label{fig:sojourn_time}
	\vspace{-2ex}
\end{figure}

As mentioned in the previous subsection, it is possible to derive the specific operational state of the machine from the available dataset. Thus, we aim to extract the production state of the machine, and, based on the state, to infer the statistical property of the network traffic (in terms of traffic pattern and distribution).
From the dataset, we could retrieve a subset of meaningful states to be used for evaluation. We will refer to these states with the following terminology: [\textit{Running}, \textit{Reentry}, \textit{Stopped}, \textit{Aborted}, \textit{Ended}]. The machine visits each of these states with a certain frequency, remains in each state for an interval of time, the so-called \textit{sojourn time}, and then transitions to a new state. For this reason, we can represent the state transition of the machine as a Markov chain. An example is given in Fig.~\ref{fig:markov}. As one can see from the figure, when the machine is in a \textit{Running} state, it can move to any of the other states, but this does not apply for all the states. For example, from an \textit{Aborted} state, it is only possible to transit to \textit{Running} and \textit{Stopped}, but not to the other two states. To give a more comprehensive understanding of those transitions, we summarize in Table~\ref{table:transition_state} the number of total transitions according to the collected dataset for any of those states. From the available dataset, the most visited states are the \textit{Running} state and the \textit{Stopped} state, since the machine most likely meets a stopping criterion after completing some task when running. While from Table~\ref{table:transition_state} one can retrieve the probability of transition among the states, another important aspect is to determine how long the machine stays in that state. This result is plotted in Fig.~\ref{fig:sojourn_time}, where the \textit{sojourn time} is shown in a logarithmic scale for readability purpose. Moreover, the logarithmic scale also helps understand the order of magnitude of such scale. As one can notice from the figure, the range of values can be very large, from few milliseconds up to minutes or even hours. Especially, the \textit{Reentry} state differs from other states, with rather a short sojourn time, while the \textit{Running} state is the one which lasts a longer time. 
By merging the transition state probability of Table~\ref{table:transition_state} together with the sojourn time probability distribution of Fig.~\ref{fig:sojourn_time}, one can empirically derive the continuous-time Markov chain to model the operational duty-cycle of the machine. 

%% file: section3.tex

To infer the network traffic process depending on the production status, first we model the production process as a semi-Markov process (SMP), then obtain the production state dependent network traffic statistics, including the distribution of packet interarrival time and its correlation with packet size. 

\subsection{Modeling the Production Process}\label{subsec:ProductionModel}

As shown in Fig. \ref{fig:markov}, the production process has five states, \emph{Running}, \emph{Reentry}, \emph{Stopped}, \emph{Aborted} and \emph{Ended}, denoted by $1$, $2$, $3$, $4$, $5$, respectively. A classical method to model such a multi-state machine is Markov process, which assumes that the sojourn time for each state follows the exponential distribution. However, in real-world applications, this assumption usually does not hold strictly. Thus, we consider to model the multi-state process of the machine as an SMP \cite{ross1996stochastic}, which relaxes the exponential assumption and allows for random sojourn time.   

To define an SMP for the production process, we consider a stochastic process $\left\{s_t\right\}_{t\geq 0}$, where $s_t\in \Ss:=\{1, 2, 3, 4, 5\}$ denotes the state at time $t$. 
\begin{itemize}
\item The \emph{transition probability} from state $i\in\Ss$ to state $j\in\Ss$ is denoted as $p_{ij}$, i.e., whenever the process is in state $i\geq 0$, the probability to enter in state $j$ is $p_{ij}$. The transition matrix is denoted as $\mP=\left[p_{ij}\right]$ with zero diagonal entries. Note that this definition of transition matrix does not include the information about the time of transition. 
\item The \emph{probability density function of jumping time}, i.e., the time needed before jumping from state $i$ to state $j$ follows a distribution denoted by $F_{ij}(t)$. 
\item The \emph{distribution of sojourn time}, i.e., the time that the process spends in state $i$ before making a transition conditioned on the next state, is given by 
\begin{equation}
H_i(t) = \sum_{j\in{\Ss}} p_{ij}F_{ij}(t).
\end{equation}
\vspace{-2ex}
\end{itemize}

From the collected data, we can extract the statistics of the transition probability (e.g., from Table \ref{table:transition_state}) and the distribution of the jumping and sojourn time (e.g., as shown in Fig. \ref{fig:sojourn_time}).

\subsection{Modeling the Production Dependent Network Traffic}\label{subsec:TrafficModel}
Given a production state, we aim to model the network traffic process conditioned to a given state by approximating the distributions of packet interarrival time. The state-of-the-art works often assume that the packet arrival follows a Poisson process, composing a limitation that the interarrival times are exponentially distributed with mean $1/\lambda$, where $\lambda$ is the arrival rate \cite{frost1994traffic}. However, a number of studies have shown that for both local-area and wide-area network traffic, the distribution of packet arrival clearly differs from exponential \cite{paxson1995wide}. 
In align with these works, we examine the traffic modeling empirically for the industrial campus network, and observe that the distribution vastly differs from Poisson process. In fact, the traffic bursts appear on a very wide range of time scales. Thus, we propose to learn the distributions of traffic arrivals with deep generative methods, such that the pre-trained models can be loaded to automatically generate production state dependent network traffic in the industrial environments. 

\subsubsection{Three Generative Models to Compare}\label{ss_sec:ModelToCompare}
\begin{figure}
\centering
\input{genModelCompare}
\caption{Comparison among three generative models, where $\vx$ and $\vx'$ are the input and reconstructed/generated samples respectively, $\vz$ is the latent vector, $y$ is a binary output which represents whether it is a real or synthesized sample, and $c$ is the condition of the production state.}
\label{fig:compareGenModel}
 \vspace{-3ex}
\end{figure}
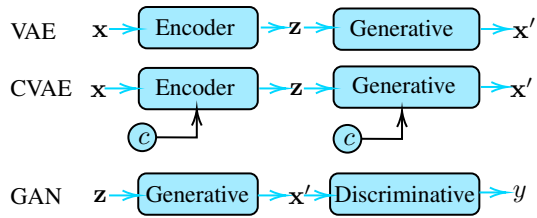

We consider the following three generative models to compare for the industrial traffic generation, as shown in Fig. \ref{fig:compareGenModel}.

{\bf Variational autoencoder (VAE)} \cite{kingma2013auto}: In VAE the input data $\vx$ is sampled from a parametrized distribution (a.k.a., the prior). The encoder maps $\vx$ to $\vz$ in a latent space, and the encoder and decoder are jointly trained to minimize a reconstruction error between the reconstructed $\vx'$ and the real input $\vx$ in terms of Kullback–Leibler (KL) divergence~\cite{kullback:divergence} between the parametric posterior and the true posterior.

{\bf Conditional variational autoencoder (CVAE)} \cite{sohn2015learning}: CVAE was developed based on the VAE architecture, with the only difference that CVAE inserts label information $c$ in the latent space to force a deterministic constrained representation (conditioned by $c$) of the learned data. In our model, the label information $c$ is the encoded categorical production state, e.g., by encoding the production state $s_t$ with one hot encoding. 

{\bf Generative adversarial network (GAN)} \cite{goodfellow2014generative}: The most popular generative model GAN simultaneously trains a generative model and a discriminative model. The generative model synthesizes samples $\vx'$ from the latent variables $\vz$, while the discriminative model differentiates between the real sample $\vx$ and the synthesized sample $\vx'$ and returns a binary output $y$ to classify both the real data and the fake data from the generator.

\subsubsection{Two Traffic Generation Schemes}\label{ss_sec:genSchemes}
Given a production state $S\in\Ss$, the objective is to generate the industrial network traffic based on the conditional distribution of packet interarrival time $T$ and packet size $L$. We compare the following two schemes.
\begin{itemize}
\item {\bf Independent Learning}: Assuming that the packet interarrival time and the packet size are independent, for each state $S\in\Ss$, we have 
$$\Pr((T, L)|s_t = S) = \Pr(T|s_t = S)\cdot \Pr(L|s_t = S).$$ 
Thus, we can learn $\Pr(T|s_t = S)$ independently from $\Pr(L|s_t = S)$ with the previously introduced generative models. 
This means, for each state $S$, we train a single model for the distribution of packet interarrival time.
\item {\bf Joint Learning}: For each state $S\in\Ss$, we learn a joint two-dimensional distribution $\Pr((T, L)|s_t = S)$. Such model captures the correlation between packet arrival time and packet size, if the correlation exists. This allows us to train one model under each production state, while the model has higher dimension of the input (taking both $T$ and $L$ into account), compared with the independent learning scheme.
\end{itemize} 

\subsection{Workflow of Generating Industrial Network Traffic}\label{ssec:Workflow}

After learning the production process as an SMP, as described in Section~\ref{subsec:ProductionModel}, and pretraining generative models for the distributions of production state dependent traffic interarrival time, we can generate the industrial traffic following the steps provided in Algorithm~\ref{alg:Workflow}.

\begin{algorithm}
\caption{Generating the industrial network traffic}\label{alg:Workflow}
\begin{algorithmic}[1]
\Require Inferred $\mP$, $\{F_{ij}\}$, and $\Pr((T, L)|S)$, $S\in\Ss$  
\Require Initial index of jump $n=0$; initial state $s_n = 1$
\Require Maximum number of state jumps $N^{(\max)}$
\While{$n \leq N^{(\max)}$}
\State Generate next state $s_{n+1}$ based on $\mP$
\State Generate jumping time based on $F_{s_n, s_{n+1}}$
 \State Before jump to $s_{n+1}$, generate $T$ and $L$ based on the pretrained geneartive models of $\Pr((T, L)|s_n)$. 
\State $n\leftarrow n+1$ when it is the time to jump to state $s_{n+1}$ 
\EndWhile
\end{algorithmic}
\end{algorithm}
\vspace{-1ex}

%% file: genModelCompare.tex
\tikzset{every picture/.style={line width=0.75pt}} 

\begin{tikzpicture}[x=0.75pt,y=0.75pt,yscale=-0.75,xscale=0.75]

\draw  [fill={rgb, 255:red, 75; green, 215; blue, 255 }  ,fill opacity=0.5 ] (123.38,56.56) .. controls (123.38,53.49) and (125.87,51) .. (128.94,51) -- (197.82,51) .. controls (200.89,51) and (203.38,53.49) .. (203.38,56.56) -- (203.38,73.25) .. controls (203.38,76.32) and (200.89,78.81) .. (197.82,78.81) -- (128.94,78.81) .. controls (125.87,78.81) and (123.38,76.32) .. (123.38,73.25) -- cycle ;
\draw  [fill={rgb, 255:red, 75; green, 215; blue, 255 }  ,fill opacity=0.5 ] (254.38,57.56) .. controls (254.38,54.49) and (256.87,52) .. (259.94,52) -- (346.82,52) .. controls (349.89,52) and (352.38,54.49) .. (352.38,57.56) -- (352.38,74.25) .. controls (352.38,77.32) and (349.89,79.81) .. (346.82,79.81) -- (259.94,79.81) .. controls (256.87,79.81) and (254.38,77.32) .. (254.38,74.25) -- cycle ;
\draw  [fill={rgb, 255:red, 75; green, 215; blue, 255 }  ,fill opacity=0.5 ] (123.38,97.56) .. controls (123.38,94.49) and (125.87,92) .. (128.94,92) -- (197.82,92) .. controls (200.89,92) and (203.38,94.49) .. (203.38,97.56) -- (203.38,114.25) .. controls (203.38,117.32) and (200.89,119.81) .. (197.82,119.81) -- (128.94,119.81) .. controls (125.87,119.81) and (123.38,117.32) .. (123.38,114.25) -- cycle ;
\draw  [fill={rgb, 255:red, 75; green, 215; blue, 255 }  ,fill opacity=0.5 ] (254.38,96.56) .. controls (254.38,93.49) and (256.87,91) .. (259.94,91) -- (346.82,91) .. controls (349.89,91) and (352.38,93.49) .. (352.38,96.56) -- (352.38,113.25) .. controls (352.38,116.32) and (349.89,118.81) .. (346.82,118.81) -- (259.94,118.81) .. controls (256.87,118.81) and (254.38,116.32) .. (254.38,113.25) -- cycle ;
\draw  [fill={rgb, 255:red, 75; green, 215; blue, 250 }  ,fill opacity=0.5 ] (116.38,138.69) .. controls (116.38,133.55) and (120.55,129.38) .. (125.69,129.38) .. controls (130.83,129.38) and (135,133.55) .. (135,138.69) .. controls (135,143.83) and (130.83,148) .. (125.69,148) .. controls (120.55,148) and (116.38,143.83) .. (116.38,138.69) -- cycle ;
\draw  [fill={rgb, 255:red, 75; green, 215; blue, 250 }  ,fill opacity=0.5 ] (254.38,139.69) .. controls (254.38,134.55) and (258.55,130.38) .. (263.69,130.38) .. controls (268.83,130.38) and (273,134.55) .. (273,139.69) .. controls (273,144.83) and (268.83,149) .. (263.69,149) .. controls (258.55,149) and (254.38,144.83) .. (254.38,139.69) -- cycle ;
\draw  [fill={rgb, 255:red, 75; green, 215; blue, 255 }  ,fill opacity=0.5 ] (123.01,168.56) .. controls (123.01,165.49) and (125.5,163) .. (128.57,163) -- (197.82,163) .. controls (200.89,163) and (203.38,165.49) .. (203.38,168.56) -- (203.38,185.25) .. controls (203.38,188.32) and (200.89,190.81) .. (197.82,190.81) -- (128.57,190.81) .. controls (125.5,190.81) and (123.01,188.32) .. (123.01,185.25) -- cycle ;
\draw  [fill={rgb, 255:red, 75; green, 215; blue, 255 }  ,fill opacity=0.5 ] (252.01,168.56) .. controls (252.01,165.49) and (254.5,163) .. (257.57,163) -- (346.82,163) .. controls (349.89,163) and (352.38,165.49) .. (352.38,168.56) -- (352.38,185.25) .. controls (352.38,188.32) and (349.89,190.81) .. (346.82,190.81) -- (257.57,190.81) .. controls (254.5,190.81) and (252.01,188.32) .. (252.01,185.25) -- cycle ;
\draw [color={rgb, 255:red, 0; green, 215; blue, 255 }  ,draw opacity=1 ][fill={rgb, 255:red, 74; green, 226; blue, 211 }  ,fill opacity=1 ]   (103.38,65.81) -- (113.38,65.81) -- (121.38,65.81) ;
\draw [shift={(123.38,65.81)}, rotate = 180] [color={rgb, 255:red, 0; green, 215; blue, 255 }  ,draw opacity=1 ][line width=0.75]    (10.93,-3.29) .. controls (6.95,-1.4) and (3.31,-0.3) .. (0,0) .. controls (3.31,0.3) and (6.95,1.4) .. (10.93,3.29)   ;
\draw [color={rgb, 255:red, 0; green, 215; blue, 255 }  ,draw opacity=1 ][fill={rgb, 255:red, 74; green, 226; blue, 211 }  ,fill opacity=1 ]   (205.38,64.81) -- (215.38,64.81) -- (219.38,64.81) ;
\draw [shift={(221.38,64.81)}, rotate = 180] [color={rgb, 255:red, 0; green, 215; blue, 255 }  ,draw opacity=1 ][line width=0.75]    (10.93,-3.29) .. controls (6.95,-1.4) and (3.31,-0.3) .. (0,0) .. controls (3.31,0.3) and (6.95,1.4) .. (10.93,3.29)   ;
\draw [color={rgb, 255:red, 0; green, 215; blue, 255 }  ,draw opacity=1 ][fill={rgb, 255:red, 74; green, 226; blue, 211 }  ,fill opacity=1 ]   (234.38,64.81) -- (247.38,64.81) -- (251.38,64.81) ;
\draw [shift={(253.38,64.81)}, rotate = 180] [color={rgb, 255:red, 0; green, 215; blue, 255 }  ,draw opacity=1 ][line width=0.75]    (10.93,-3.29) .. controls (6.95,-1.4) and (3.31,-0.3) .. (0,0) .. controls (3.31,0.3) and (6.95,1.4) .. (10.93,3.29)   ;
\draw [color={rgb, 255:red, 0; green, 215; blue, 255 }  ,draw opacity=1 ][fill={rgb, 255:red, 74; green, 226; blue, 211 }  ,fill opacity=1 ]   (353.38,65.81) -- (366.38,65.81) -- (370.38,65.81) ;
\draw [shift={(372.38,65.81)}, rotate = 180] [color={rgb, 255:red, 0; green, 215; blue, 255 }  ,draw opacity=1 ][line width=0.75]    (10.93,-3.29) .. controls (6.95,-1.4) and (3.31,-0.3) .. (0,0) .. controls (3.31,0.3) and (6.95,1.4) .. (10.93,3.29)   ;
\draw [color={rgb, 255:red, 0; green, 215; blue, 255 }  ,draw opacity=1 ][fill={rgb, 255:red, 74; green, 226; blue, 211 }  ,fill opacity=1 ]   (100.38,105.81) -- (110.38,105.81) -- (119.38,105.81) ;
\draw [shift={(121.38,105.81)}, rotate = 180] [color={rgb, 255:red, 0; green, 215; blue, 255 }  ,draw opacity=1 ][line width=0.75]    (10.93,-3.29) .. controls (6.95,-1.4) and (3.31,-0.3) .. (0,0) .. controls (3.31,0.3) and (6.95,1.4) .. (10.93,3.29)   ;
\draw [color={rgb, 255:red, 0; green, 215; blue, 255 }  ,draw opacity=1 ][fill={rgb, 255:red, 74; green, 226; blue, 211 }  ,fill opacity=1 ]   (103.38,177.81) -- (113.38,177.81) -- (121.38,177.81) ;
\draw [shift={(123.38,177.81)}, rotate = 180] [color={rgb, 255:red, 0; green, 215; blue, 255 }  ,draw opacity=1 ][line width=0.75]    (10.93,-3.29) .. controls (6.95,-1.4) and (3.31,-0.3) .. (0,0) .. controls (3.31,0.3) and (6.95,1.4) .. (10.93,3.29)   ;
\draw [color={rgb, 255:red, 0; green, 215; blue, 255 }  ,draw opacity=1 ][fill={rgb, 255:red, 74; green, 226; blue, 211 }  ,fill opacity=1 ]   (204.38,105.81) -- (214.38,105.81) -- (220.38,105.81) ;
\draw [shift={(222.38,105.81)}, rotate = 180] [color={rgb, 255:red, 0; green, 215; blue, 255 }  ,draw opacity=1 ][line width=0.75]    (10.93,-3.29) .. controls (6.95,-1.4) and (3.31,-0.3) .. (0,0) .. controls (3.31,0.3) and (6.95,1.4) .. (10.93,3.29)   ;
\draw [color={rgb, 255:red, 0; green, 215; blue, 255 }  ,draw opacity=1 ][fill={rgb, 255:red, 74; green, 226; blue, 211 }  ,fill opacity=1 ]   (234.38,105.81) -- (247.38,105.81) -- (251.38,105.81) ;
\draw [shift={(253.38,105.81)}, rotate = 180] [color={rgb, 255:red, 0; green, 215; blue, 255 }  ,draw opacity=1 ][line width=0.75]    (10.93,-3.29) .. controls (6.95,-1.4) and (3.31,-0.3) .. (0,0) .. controls (3.31,0.3) and (6.95,1.4) .. (10.93,3.29)   ;
\draw [color={rgb, 255:red, 0; green, 215; blue, 255 }  ,draw opacity=1 ][fill={rgb, 255:red, 74; green, 226; blue, 211 }  ,fill opacity=1 ]   (203.38,177.81) -- (216.38,177.81) -- (220.38,177.81) ;
\draw [shift={(222.38,177.81)}, rotate = 180] [color={rgb, 255:red, 0; green, 215; blue, 255 }  ,draw opacity=1 ][line width=0.75]    (10.93,-3.29) .. controls (6.95,-1.4) and (3.31,-0.3) .. (0,0) .. controls (3.31,0.3) and (6.95,1.4) .. (10.93,3.29)   ;
\draw [color={rgb, 255:red, 0; green, 215; blue, 255 }  ,draw opacity=1 ][fill={rgb, 255:red, 74; green, 226; blue, 211 }  ,fill opacity=1 ]   (234.38,177.81) -- (246.38,177.81) -- (250.38,177.81) ;
\draw [shift={(252.38,177.81)}, rotate = 180] [color={rgb, 255:red, 0; green, 215; blue, 255 }  ,draw opacity=1 ][line width=0.75]    (10.93,-3.29) .. controls (6.95,-1.4) and (3.31,-0.3) .. (0,0) .. controls (3.31,0.3) and (6.95,1.4) .. (10.93,3.29)   ;
\draw [color={rgb, 255:red, 0; green, 215; blue, 255 }  ,draw opacity=1 ][fill={rgb, 255:red, 74; green, 226; blue, 211 }  ,fill opacity=1 ]   (353.38,104.81) -- (365.38,104.81) -- (369.38,104.81) ;
\draw [shift={(371.38,104.81)}, rotate = 180] [color={rgb, 255:red, 0; green, 215; blue, 255 }  ,draw opacity=1 ][line width=0.75]    (10.93,-3.29) .. controls (6.95,-1.4) and (3.31,-0.3) .. (0,0) .. controls (3.31,0.3) and (6.95,1.4) .. (10.93,3.29)   ;
\draw [color={rgb, 255:red, 0; green, 215; blue, 255 }  ,draw opacity=1 ][fill={rgb, 255:red, 74; green, 226; blue, 211 }  ,fill opacity=1 ]   (352.38,175.81) -- (364.38,175.81) -- (368.38,175.81) ;
\draw [shift={(370.38,175.81)}, rotate = 180] [color={rgb, 255:red, 0; green, 215; blue, 255 }  ,draw opacity=1 ][line width=0.75]    (10.93,-3.29) .. controls (6.95,-1.4) and (3.31,-0.3) .. (0,0) .. controls (3.31,0.3) and (6.95,1.4) .. (10.93,3.29)   ;
\draw    (135,138.69) -- (162.38,138.81) -- (162.38,121.81) ;
\draw [shift={(162.38,119.81)}, rotate = 90] [color={rgb, 255:red, 0; green, 0; blue, 0 }  ][line width=0.75]    (10.93,-3.29) .. controls (6.95,-1.4) and (3.31,-0.3) .. (0,0) .. controls (3.31,0.3) and (6.95,1.4) .. (10.93,3.29)   ;
\draw    (273,139.69) -- (300.38,139.81) -- (300.38,122.81) ;
\draw [shift={(300.38,120.81)}, rotate = 90] [color={rgb, 255:red, 0; green, 0; blue, 0 }  ][line width=0.75]    (10.93,-3.29) .. controls (6.95,-1.4) and (3.31,-0.3) .. (0,0) .. controls (3.31,0.3) and (6.95,1.4) .. (10.93,3.29)   ;

\draw (35,60) node [anchor=north west][inner sep=0.75pt]   [align=left] {{\small VAE}};
\draw (132.56,57) node [anchor=north west][inner sep=0.75pt]   [align=left] {{\small Encoder}};
\draw (263.38,57.56) node [anchor=north west][inner sep=0.75pt]   [align=left] {{\small Generative}};
\draw (90,61.4) node [anchor=north west][inner sep=0.75pt]    {$\vx$};
\draw (222,59.4) node [anchor=north west][inner sep=0.75pt]    {$\vz$};
\draw (373,56.4) node [anchor=north west][inner sep=0.75pt]    {$\vx'$};
\draw (35,98) node [anchor=north west][inner sep=0.75pt]   [align=left] {{\small CVAE}};
\draw (132.56,97) node [anchor=north west][inner sep=0.75pt]   [align=left] {{\small Encoder}};
\draw (264.94,96) node [anchor=north west][inner sep=0.75pt]   [align=left] {{\small Generative}};
\draw (89,101.4) node [anchor=north west][inner sep=0.75pt]    {$\vx$};
\draw (223,100.4) node [anchor=north west][inner sep=0.75pt]    {$\vz$};
\draw (371,96.4) node [anchor=north west][inner sep=0.75pt]    {$\vx'$};
\draw (121,133.4) node [anchor=north west][inner sep=0.75pt]    {$c$};
\draw (259,134.4) node [anchor=north west][inner sep=0.75pt]    {$c$};
\draw (35,171) node [anchor=north west][inner sep=0.75pt]   [align=left] {{\small GAN}};
\draw (126.01,169.56) node [anchor=north west][inner sep=0.75pt]   [align=left] {{\small Generative}};
\draw (253, 169) node [anchor=north west][inner sep=0.75pt]   [align=left] {{\small Discriminative}};
\draw (91,172.4) node [anchor=north west][inner sep=0.75pt]    {$\vz$};
\draw (222,168.4) node [anchor=north west][inner sep=0.75pt]    {$\vx'$};
\draw (373,169.4) node [anchor=north west][inner sep=0.75pt]    {$y$};

\end{tikzpicture}

%% file: section4.tex
\begin{figure*}
	\centering
	\begin{subfigure}[b]{0.245\textwidth}
		\includegraphics[width=\textwidth]{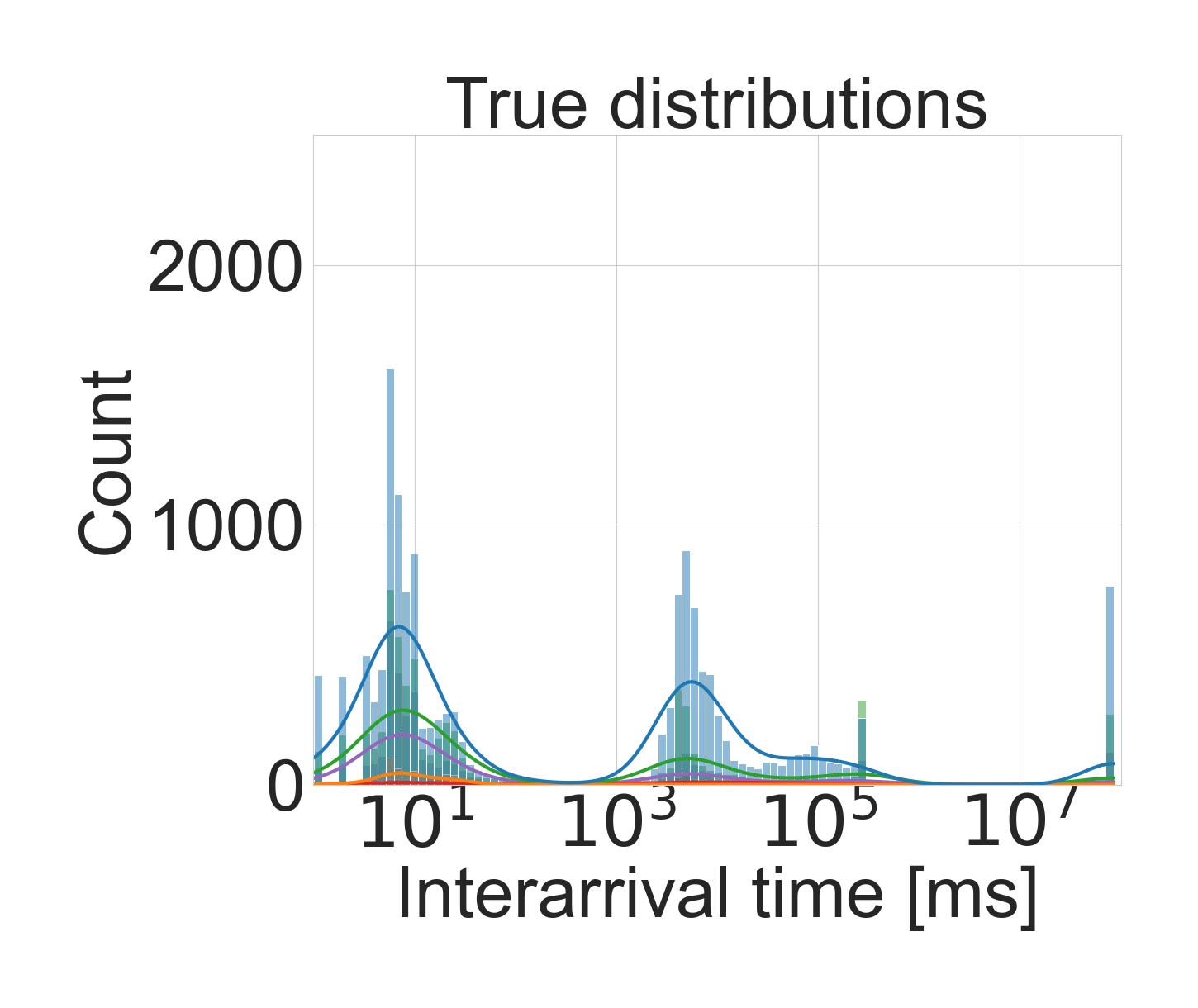}
        \caption{}
		\label{fig:true_1D}
	\end{subfigure}
	\hspace{-2.0ex}
    \begin{subfigure}[b]{0.245\textwidth}
	\centering
		\includegraphics[width=\textwidth]{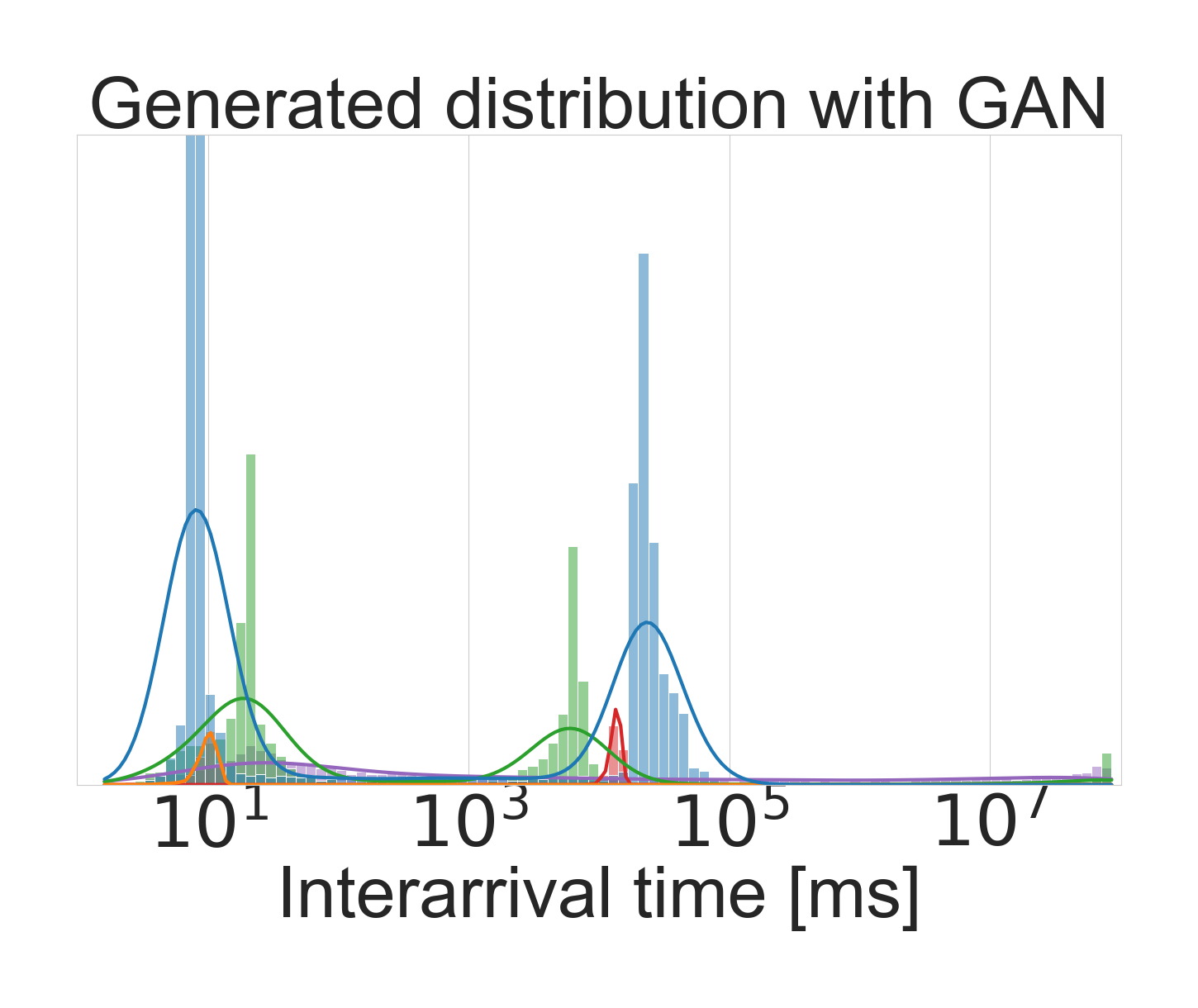}
        \caption{}
			\label{fig:gan_1D}
	\end{subfigure}
	\hspace{-2.0ex}
	\begin{subfigure}[b]{0.245\textwidth}
		\includegraphics[width=\textwidth]{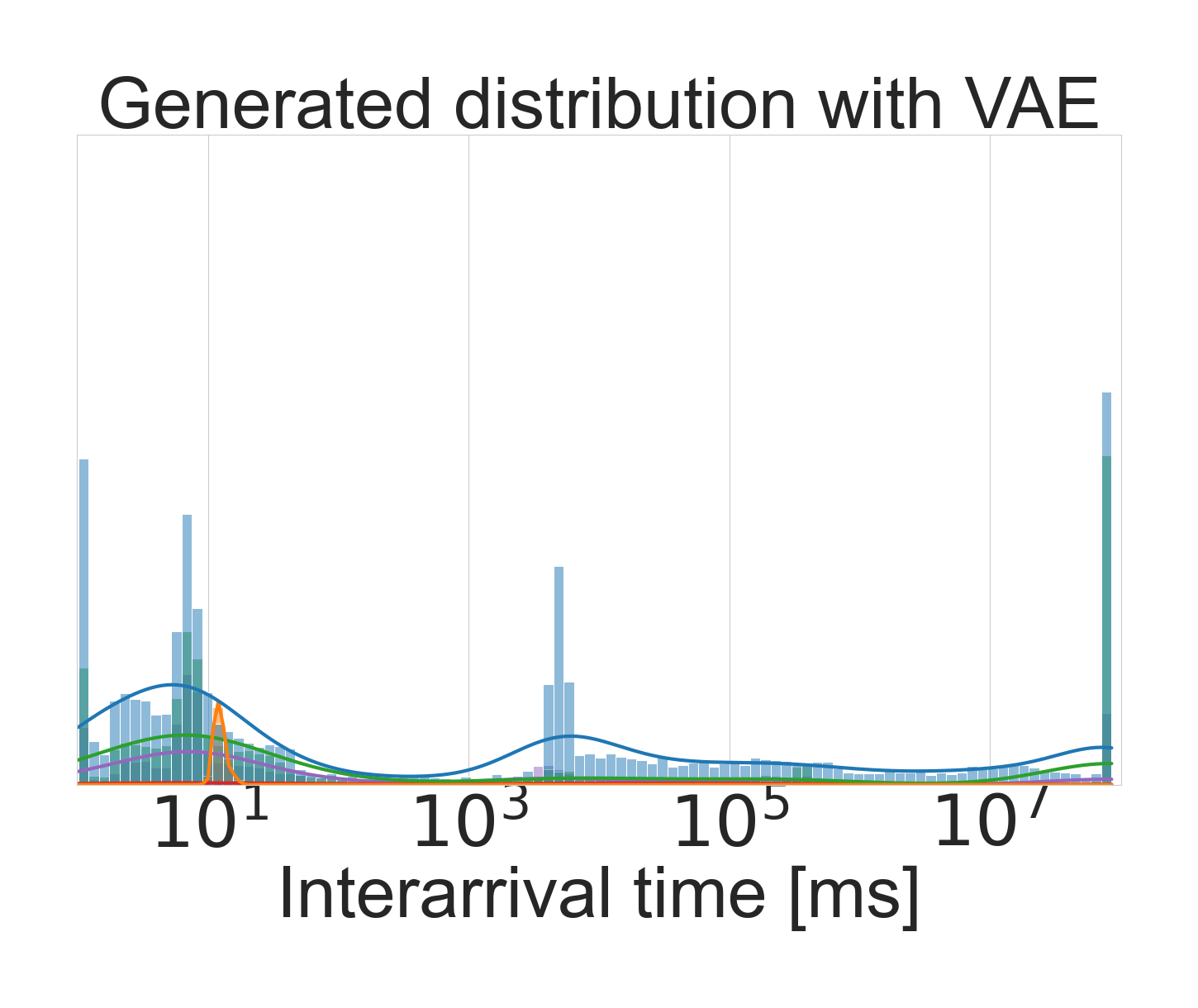}
        \caption{}
		\label{fig:vae_1D}
	\end{subfigure}
	\hspace{-2.0ex}
	\begin{subfigure}[b]{0.245\textwidth}
		\includegraphics[width=\textwidth]{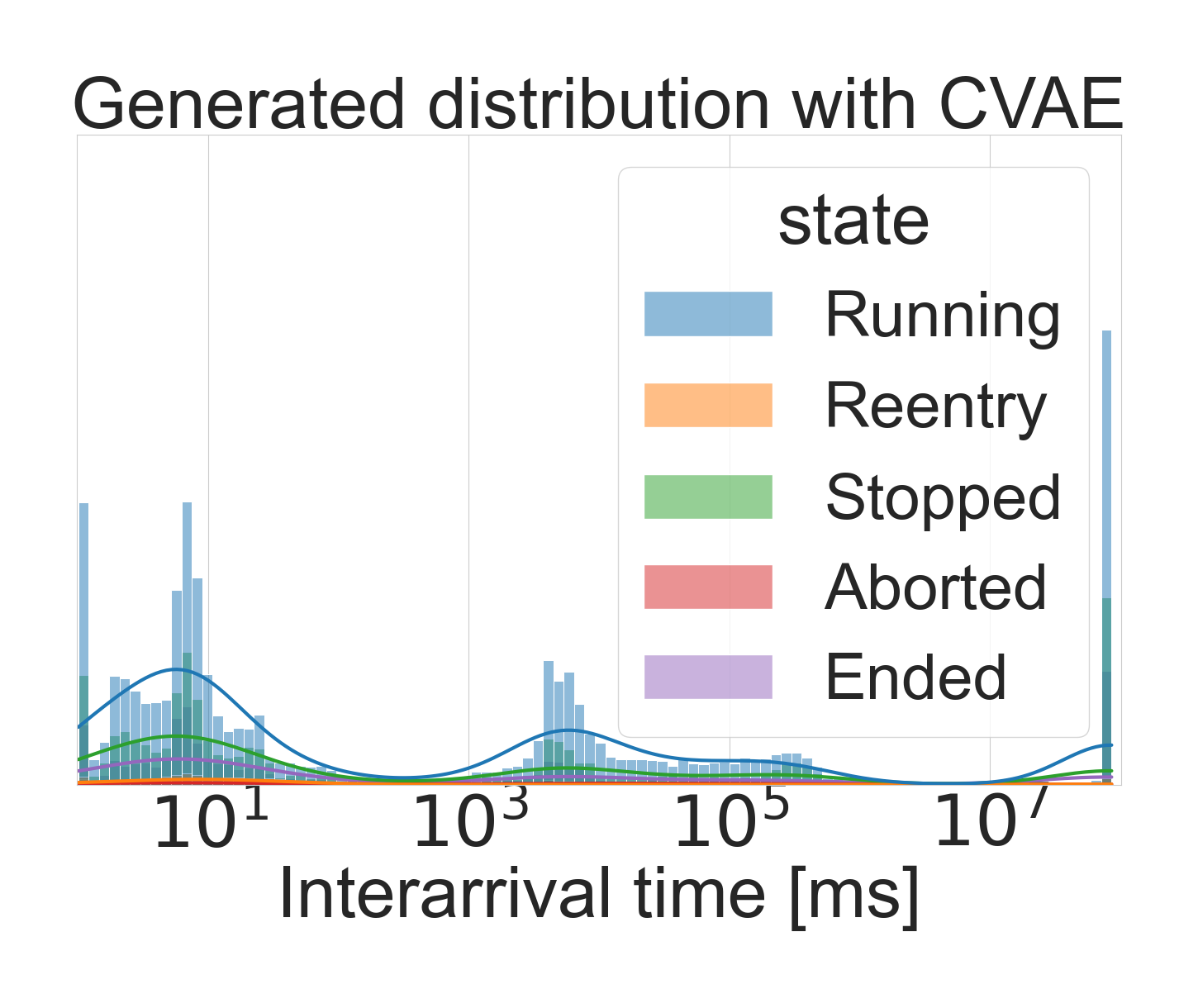}
        \caption{}
		\label{fig:cvae_1D}
	\end{subfigure}
	\caption{Comparison of distributions for all states}
	\label{fig:distribution_1D}
	\vspace{-0.3cm}
\end{figure*}

In this section we show the performance of the generative models discussed in Section~\ref{sec:3} applied to the dataset illustrated in Section~\ref{sec:2}. In particular, we consider two setups for the generation of the traffic:
\begin{enumerate}
	\item We use VAE and GAN architectures to learn traffic distribution for each production state of the machine independently and compare it with the empirical distribution of the real data. This results in multiple per-state models, which is helpful for the evaluation of the behavior of the industrial environment in the specific states.
	\item We use CVAE architecture to learn the state-dependent traffic distribution jointly for all states by adding the production state as a priori information. This leads to a single model that generates the distribution during the whole operational duty-cycle of the machine. 
\end{enumerate}
The two approaches give flexibility to apply the generative models in different contexts, depending on the need. 
In particular, we apply i) the GAN and VAE on a per-state dataset, where each dataset is extracted by the original dataset and filtered according to the production state of the machine. In contrast, for CVAE we consider ii) the production state of the machine as a priori information given to the neural networks as the label condition $c$. The main difference between the two approaches is that in i) we obtain multiple models, one for each production state, while in ii) we have a single model that can generate the per-state traffic patterns for any production state.
For all three generative models, we consider the same neural network architecture and hyper-parameter tuning. The models of encoder and decoder for VAE and CVAE, as well as generator and discriminator for the GAN, are all built up with 5 hidden layers, with the number of neurons (32, 64, 32, 16, 32). In all schemes we adopt a learning rate of 0.001, and the loss functions are built upon the binary cross entropy, with the addition of the Kullback-Leibler (KL) loss for the VAE and CVAE. The latent dimensions for interarrival time distribution only and for the joint distribution of interarrival time and packet size are $2$ and $4$, respectively. The training is performed on the $70\%$ of the whole dataset, which is shuffled and normalized after being rescaled on a logarithmic scale, with batch size of $32$, for a total of $500$ epochs. The numbers of samples for each state are not homogeneous, resulting in a total of ca. $15000$ of samples for \textit{Running} state and only ca. $500$ for \textit{Aborted} and \textit{Reentry} states. For such reason, the testing is performed on variable number of samples for each state, but with a sufficient number to draw meaningful distributions.

Below, we show the experimental results when generating
\begin{itemize}
	\item the distribution of interarrival time of packets only;
	\item the joint distribution of interarrival time and packet size.
\end{itemize}
\subsection{Distribution of interarrival time of packets}
One of the most important features for traffic estimation and generation is the interarrival time between packets. This parameter refers to the interval time between two consecutive transmissions and it is typically varying depending on the specific applications generating the data traffic. We focus on the interarrival time since it is also a fundamental unit of measure for critical applications, where the network must ensure the correct delivery of the packets in the range of few milliseconds. Therefore, understanding the interarrival time of packets gives an important information for scheduling operations and how to handle priorities among packets. 
We analyze the different characteristics of the interarrival time for each state, as defined in Section~\ref{sec:2}, and compare the generated distributions with the empirical distribution computed from the original samples. 
Namely, we compare the original distribution in Fig.~\ref{fig:true_1D} with the ones generated with GAN, VAE and CVAE in Figs.~\ref{fig:gan_1D}, ~\ref{fig:vae_1D} and ~\ref{fig:cvae_1D}, respectively.
In particular, the shaded bars show the histograms obtained from the distributions, while the solid lines show the approximation of the continuous distributions generated by the seaborn modules of \textit{histplot}. For each figure, we compare the generated distribution with the empirical distribution to analyze the properties and similarities between the plotted samples. Moreover, the x-axis is plotted in logarithmic scale for the ease of visualization, since the range of the samples is quite large, from few milliseconds till minutes of interarrival time. However, the logarithmic scale gives a good understanding of the order of magnitude, showing that, for example, in the \textit{Running} state it is possible to identify three main peaks, the first in the order of tens of milliseconds, the second in the order of tens of seconds, while the third one in the order of minutes. When comparing the results for each state, one can appreciate that VAE and CVAE generate, in general, distributions very close to the real data. In contrast, GAN seems unable to intercept all the peaks of the original distributions, in particular when observing the samples of interarrival times in the order of minutes. By looking at individual states, one can also notice how some states cannot be retrieved properly, both by GAN and VAE, due to the limited amount of data for the training. This is particularly true for the \textit{Aborted} and \textit{Reentry} states, which are the states less visited by the machine and, therefore, produce limited amount of samples. However, CVAE also gives a good approximation for those states, resulting in good accuracy for each state.
To give a more comprehensive comparison among the proposed models, we also compare the KL divergence for each state in Table~\ref{table:KL_divergence}. The values considering the interarrival time only are referred to as VAE 1D, CVAE 1D, GAN 1D. As already shown from the plotted distributions, the CVAE obtains in general better results, providing similar values for each state, differently from VAE and GAN. However, VAE has comparable (if not better) performance for most of the states with enough samples for the training. 
Finally, the CVAE is able to generalize in a single model the distributions for all states, given also the high similarity of those distributions among some states.




\begin{figure*}
	\centering
	\begin{subfigure}[b]{0.245\textwidth}
		\includegraphics[width=\textwidth]{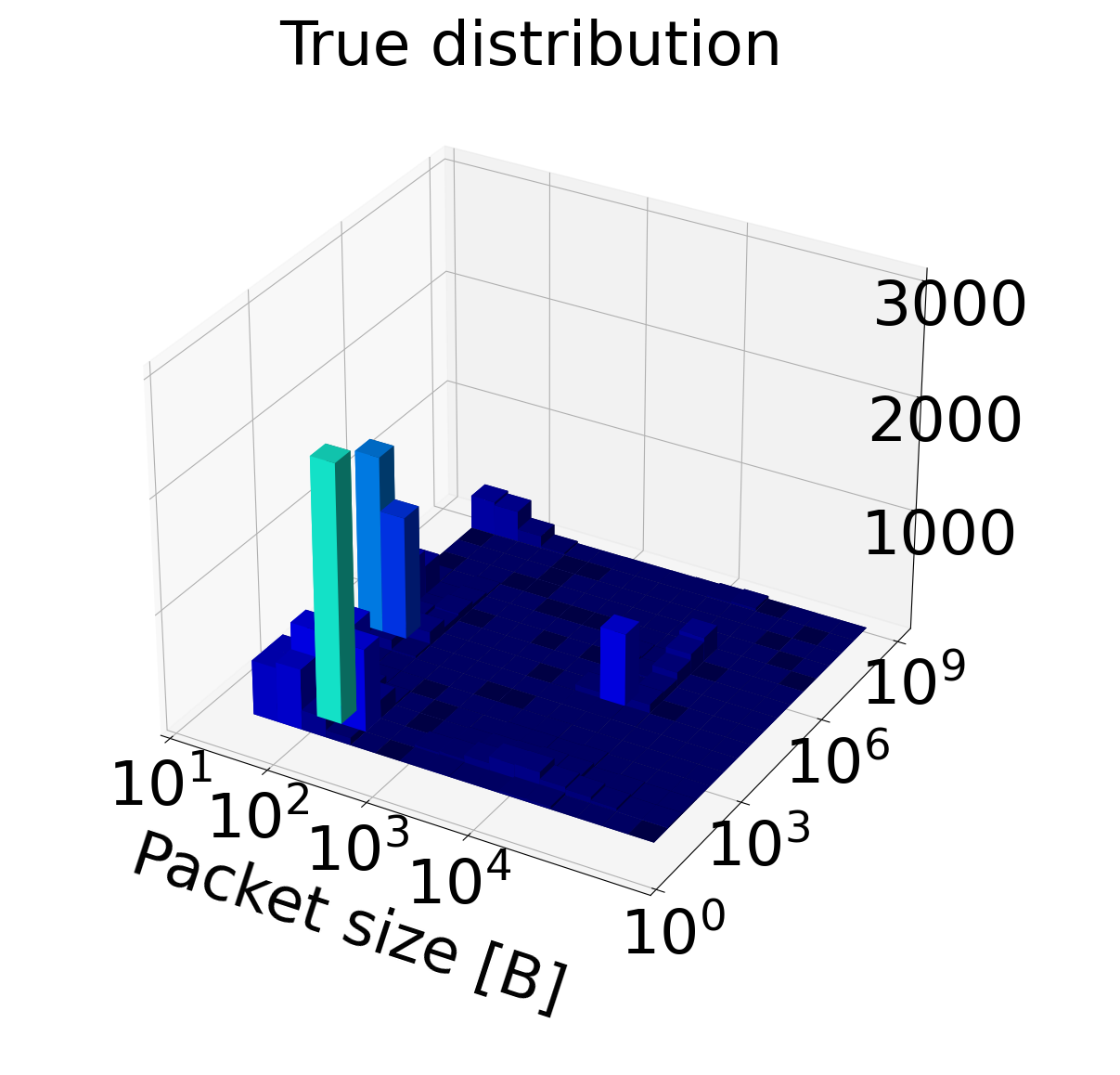}
        \caption{}
		\label{fig:true_2D_run}
	\end{subfigure}
	\hspace{-2.0ex}
	\begin{subfigure}[b]{0.245\textwidth}
	\centering
		\includegraphics[width=\textwidth]{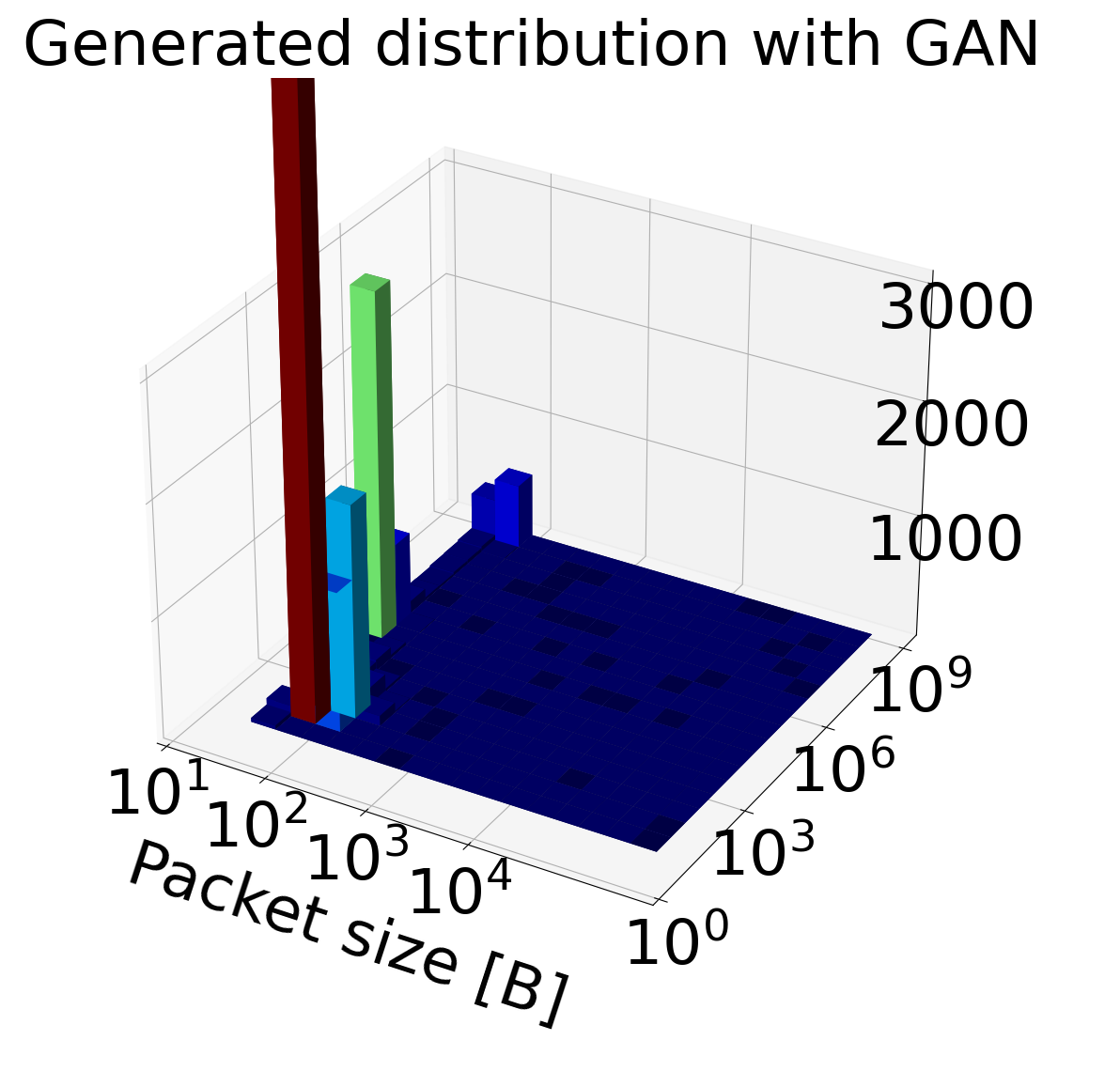}
        \caption{}
		\label{fig:gan_2D_run}
	\end{subfigure}
	\hspace{-2.0ex}
	\begin{subfigure}[b]{0.245\textwidth}
		\includegraphics[width=\textwidth]{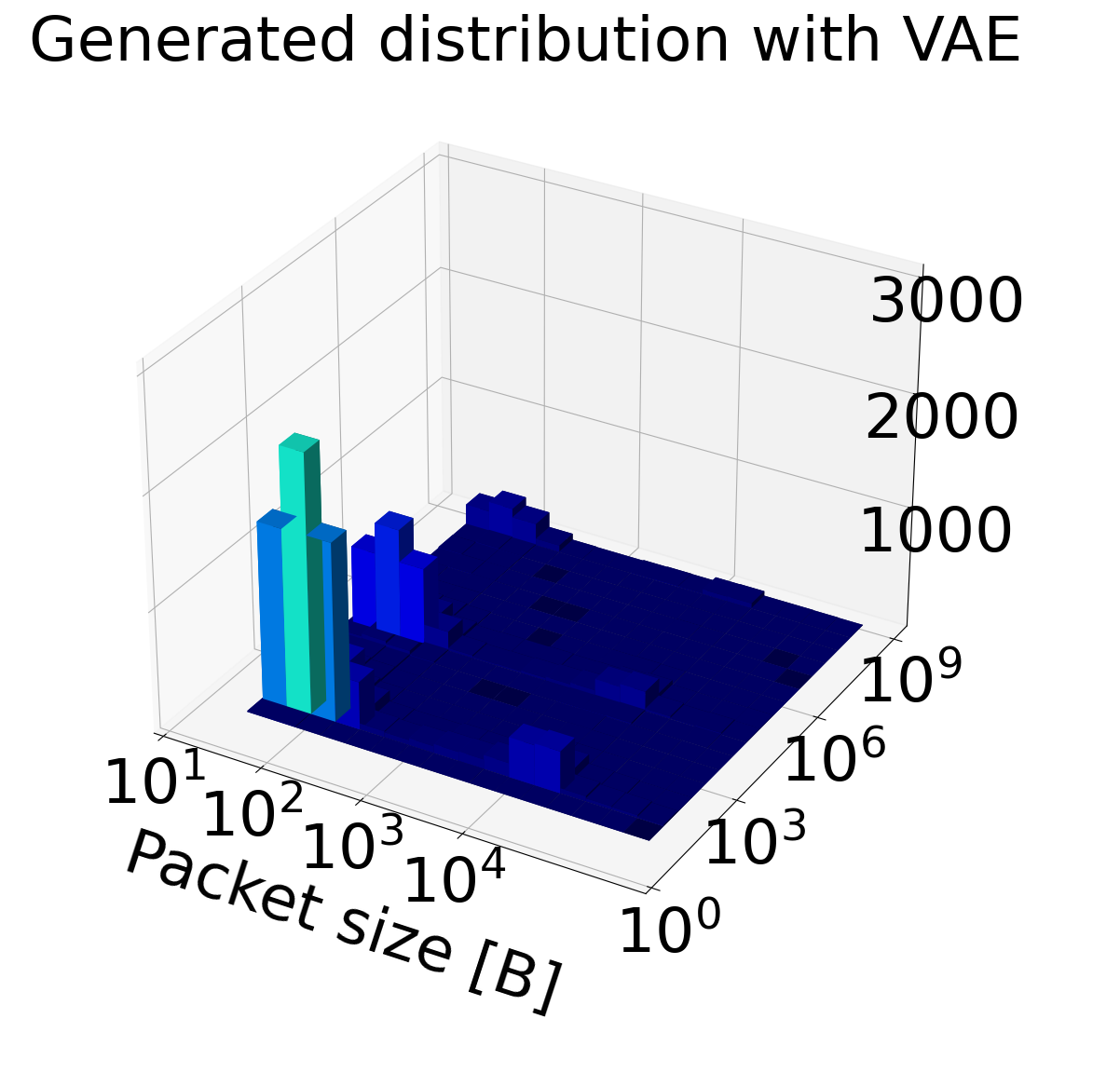}
        \caption{}
		\label{fig:vae_2D_run}
	\end{subfigure}
	\hspace{-2.0ex}
	\begin{subfigure}[b]{0.245\textwidth}
		\includegraphics[width=\textwidth]{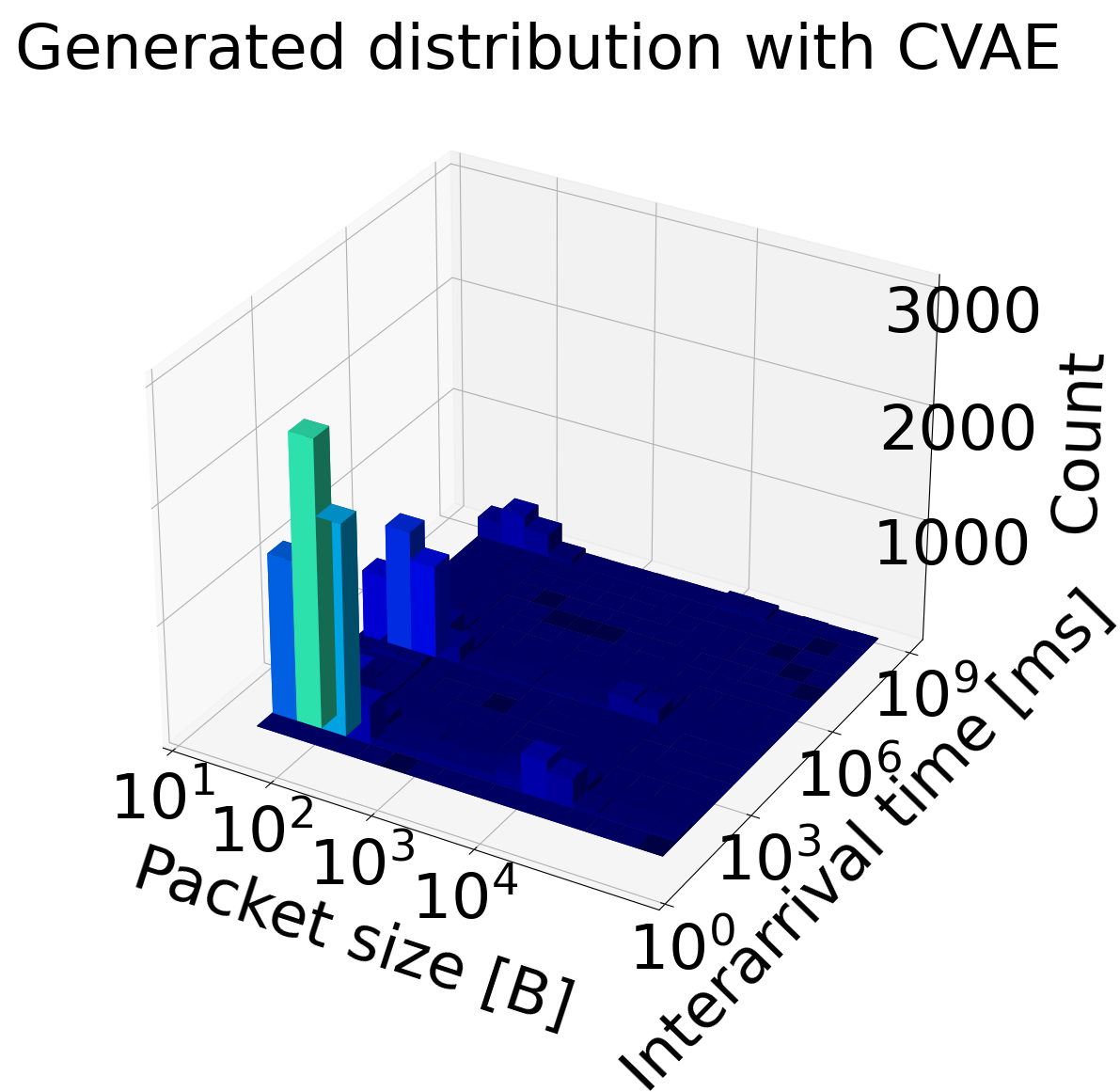}
        \caption{}
		\label{fig:cvae_2D_run}
	\end{subfigure}
	\caption{Comparison of the joint 2D distribution for the \textit{Running} state}
	\label{fig:Running_2D}
	\vspace{-0.4cm}
\end{figure*}

\subsection{Joint distribution of interarrival time and packet size}

In this section, we analyze the joint distribution of interarrival time and packet size by comparing the results of the 2D distributions generated by the three models with the empirical distribution. For the sake of space and clarity, we only show a 3D plot of the 2D distribution for the \textit{Running} state in Fig.~\ref{fig:Running_2D}, 
 while referring at the KL divergence values in Table~\ref{table:KL_divergence}, computed for the interarrival time distribution only, for generic comparison and discussion. 
The plots in the figure show the bars in the different colors to further highlight the difference in number of counts in the obtained histogram. We apply the logarithmic scale at both x-axis and y-axis for the interrarrival time (in ms) and packet size (in bytes),  respectively. One can notice that most of the packets for \textit{Running} state are transmitted with short interarrival time (of few milliseconds) and have small packet sizes, while only few transmissions are established with large packet sizes. 
As for the 1D distribution, the VAE and CVAE outperforms GAN, showing more similarities in the distribution, confirmed by the lower values in the KL divergence in Table~\ref{table:KL_divergence} (comparing the values of VAE 2D, CVAE 2D, GAN 2D). By comparing the values in Table~\ref{table:KL_divergence}, one can deduce that interarrival time and packet size are not strongly correlated, given that the performance of the joint distribution do not outperform the one obtained with the interarrival time only. We remark that the scope of this paper is to give insights about the traffic pattern in industrial network and to share the generative models discussed for reproducing the realistic environments. The provided generative models can be used by the entire research community to build more realistic simulators and emulators for the industrial 5G use cases. The pre-trained model can be found in~\cite{git:models2022}.

\begin{table}
\vspace{.4cm}
	\centering
	\begin{tabular}{ | c | c | c | c | c | c |}
		\hline
			  					& \textit{Running}       & \textit{Reentry}  	& \textit{Stopped}	& \textit{Aborted}	& \textit{Ended} 		\\ \hline
		VAE 1D						& $0.40$   		& $4.27$			& $0.52$   		& $0.50$			& $\mathbf{0.34}$	\\ \hline
		CVAE 1D						& $\mathbf{0.26}$   	& $0.38$			& $\mathbf{0.34}$   	& $\mathbf{0.47}$	& $0.36$			\\ \hline
		GAN 1D  						& $2.54$      		& $2.25$			& $1.23$   		& $4.43$			& $1.33$	     		\\ \hline
		VAE 2D        					& $0.33$ 			& $\mathbf{0.35}$ 	& $0.49$ 			& $0.78$			& $0.35$			\\ \hline
		CVAE 2D        					& $0.28$ 	 		& $0.95$ 			& $0.36$ 			& $0.69$			& $0.38$			\\ \hline
		GAN 2D        					& $1.02$  			& $2.28$ 			& $2.08$   		& $4.43$			& $1.74$	 		\\ \hline

	\end{tabular}
	\caption{Model comparison based on KL divergence}
	\label{table:KL_divergence}
	\vspace{-.4cm}
\end{table}

%% file: conclusion.tex
In this paper we analyze the traffic patterns of a laser cutting machine deployed in a Trumpf factory. From the collected dataset, we can identify the different production states of the machine and extract the production state dependent statistics of the industrial network traffic. Specifically, we show that the production state process can be represented as a semi-Markov process, by modeling the transition probabilities and the sojourn time for each state. Moreover, we propose different architectures for generating traffic models that can mimic the statistics of the production dataset. VAE and GAN are considered for generating per-state traffic models, while CVAE is considered as a generic model to generate the traffic patterns for every state, by adding the conditional information of the state to the generator.
We compare the performance of the different architectures and observe that CVAE reproduces samples closer to the empirical distribution computed from real data when considering only the statistics of the interarrival time, while VAE outperforms the other models for the joint distribution of interarrival time and packet size. The obtained models are also made publicly available to allow researchers to reproduce more realistic industrial environments for their research activities. 
The integration of those models in an industrial network simulator will be part of the future works.

%% file: acknowledgement.tex
This work was supported by the German Federal Ministry of Education and Research (BMBF) project KICK.